# Evolutionary Power-Aware Routing in VANETs using Monte-Carlo Simulation

Jamal Toutouh

University of Malaga

Spain

Sergio Nesmachnow

Universidad de la Republica

Uruguay

Enrique Alba

University of Malaga

Spain

# Evolutionary Power-Aware Routing in VANETs using Monte-Carlo Simulation

Jamal Toutouh
University of Malaga
Spain
Email: jamal@lcc.uma.es

Sergio Nesmachnow
Universidad de la Republica
Uruguay
Email: sergion@fing.edu.uy

Enrique Alba
University of Malaga
Spain
Email: eat@lcc.uma.es

*Abstract*—This work addresses the reduction of power consumption of the AODV routing protocol in vehicular networks as an optimization problem. Nowadays, network designers focus on energy-aware communication protocols, specially to deploy wireless networks. Here, we introduce an automatic method to search for energy-efficient AODV configurations by using an evolutionary algorithm and parallel Monte-Carlo simulations to improve the accuracy of the evaluation of tentative solutions. The experimental results demonstrate that significant power consumption improvements over the standard configuration can be attained, with no noteworthy loss in the quality of service.

*Index Terms*—Vehicular Ad hoc Networks, Energy-efficient, AODV, Differential Evolution, Monte-Carlo Simulation

## I. INTRODUCTION

The current improvements in wireless communication and information technologies involve important advances in vehicular ad hoc networks (VANETs). VANETs are generated spontaneously to exchange information between nearby vehicles, roadside infrastructure, sensors, and pedestrian devices. Powerful intelligent transportation system applications capable of gathering, processing, and distributing up-to-minute information for road users are developed using these technologies [1]. In VANETs, nodes communicate directly with their neighbors and require decentralized routing protocols to propagate the information. The use of WiFi technologies with limited coverage and the intrinsic high node mobility make difficult the routing, due to frequent topology changes and network fragmentation. Many routing protocols have appeared for VANETs, but existing mobile ad hoc networks (MANET) protocols are also employed [2]. In this work, we focus on Ad hoc On Demand Vector (AODV), a well-known MANET routing protocol that has been used in VANETs [3], [4].

VANETs involve devices fed by limited power sources such as sensors, pedestrian devices and smartphones, traffic signs, etc. The power consumption of wireless communications and computations becomes a major concern in VANETs, and the use of energy-aware approaches, as power-aware network architectures and protocols design, are highly desirable [5], [6]. Thus, network capabilities are maintained by extending the battery life of the devices. Most of the power aware techniques used in VANETs are based on those employed in wireless sensor networks (WSN) and MANETs [7], [8]; e.g., the power transmission adjustment [6], the power save mechanism (PSM), and the use of the previous two together [7].

In this paper, we deal with the energy efficiency in the AODV routing protocol. The node power consumption is affected by the type of routing protocol in two different ways: the routing workload generated by the protocol impacts on the amount of energy used for exchanging routing control messages, and the generated routing paths affect the nodes consumption when forwarding packets.

Interesting power-aware techniques have been applied to existing routing protocols [9]. The methods to reduce the AODV power consumption include *Power-Aware Routing* where the routing path is computed by minimizing the power needed to transfer the data; *Local Energy-Aware Routing*, where the battery levels on each node are used to decide whether or not to participate in the route discovering process; and *Lifetime Prediction Routing*, where the routing paths are composed with the terminals with more remaining power [10].

In this article, we study the application of the Differential Evolution (DE) metaheuristic to find power-aware AODV configurations. Some works have used metaheuristic algorithms to analyze vehicular communication protocols, mainly to optimize the quality of service (QoS) of VDTP [11], AODV [12], and OLSR [13] protocols. Our previous work [14] studied efficient power-aware configurations for OLSR in VANETs.

In our approach, the optimization process is guided by evaluating tentative solutions (protocol configurations) by means of VANET simulations. To obtain accurate simulations, it is necessary to employ signal propagation models reflecting real world interactions, as the stochastic Nakagami propagation model [15]. Due to the stochastic process involved, the simulation of the same scenario returns different results for the metric evaluated. Unlike the previous works—that use just one simulation to evaluate a solution—we perform a number of simulations in parallel, applying a Monte-Carlo method [16] for simulation. Thereby, the search for optimized solutions is improved because it is guided by more accurate evaluations.

Summarizing, the main contribution of the manuscript are: i) to introduce an optimization methodology for power-aware AODV in VANETs, using the DE algorithm and the Monte Carlo method to improve the evaluation of tentative solutions, by performing a number of VANET simulations in parallel; and ii) to improve the power-efficiency of AODV in VANETs in comparison with the Request For Coments (RFC) 3561 predefined configuration [17].

The paper is organized as follows. The next section introduces the energy aware routing problem in AODV. Section III describes the DE method used in this work, and Section IV presents the methodology and the implementation details of the proposed DE for power-aware AODV in VANETs. The experimental evaluation of the proposed method is presented in Section V. Finally, Section VI presents the conclusions of the research and formulates the main lines for future work.

## II. POWER-AWARE AODV ROUTING IN VANETs

Routing protocols determine the way that the devices exchange information, by establishing and maintaining the routes, making decision for forwarding packets, and recovering from routing failures. The power consumption of each node is influenced by these protocols in the routing table operations and the packet forwarding. In this work, we deal with power-aware routing in VANETs when using AODV protocol. This section presents AODV and the power consumption problem.

### A. Ad hoc On Demand Vector

Ad hoc On Demand Vector (AODV) [17] is a reactive routing protocol for ad hoc networks designed to overcome the overhead problem of its precedent, Destination Sequence Distance Vector (DSDV). As a reactive protocol, AODV determines the routes when a source node has data traffic to send, and it maintains just the paths that are currently in use. Thus, it reduces the routing overload generated by proactive routing protocols to maintain the routing paths at any time.

AODV reduces the routing network load by broadcasting route discovery mechanism and by dynamically updating routing information at each intermediate node in the path. A source node starts the route discovery process by broadcasting a Routing Request (RREQ) packet, in order to compute the routing path to a particular destination. Neighbor nodes which do not know an active route for the requested destination forward the RREQ packet to their neighbors, until an active route is found or the maximum number of hops is reached. When an intermediate node knows an active route to the requested destination, it sends a Route Replay (RREP) packet back to the source node in unicast mode. Finally, the source node receives the RREP packet including the route information. If the source node does not receive any RREP packet within a given time, it assumes that route is not available.

The operation and performance of AODV is significantly influenced by the value of its control parameters. These parameters are principally five timers and six counters. The AODV RFC 3561 suggests a generic MANET parameterization (see Table I) that has been also used in vehicular networks [3], [4]. However, alternative configurations have outperformed the RFC configurations in VANETs, regarding QoS [12].

### B. AODV Power-Aware Parameter Tuning

The primary objectives of mobile routing protocols are to maximize the network throughput, the energy efficiency, and the network (terminals) life time. The power consumption is highly conditioned by the network overhead, therefore protocols that generate lower routing load typically consumes less power. Proactive routing protocols, e.g., OLSR and DSDV, are more energy-efficient because they produce lower overhead in VANETs and reactive approaches offer better throughput in such networks [18]. Therefore, our approach aims at improving the energy efficiency of AODV without adversely affecting the proper delivery of packets.

TABLE I
SET OF AODV PARAMETERS AND AODV RFC 3561 SPECIFIED VALUES.

| parameter | RFC value | type | range |
|---|---|---|---|
| HELLO_INTERVAL | 1.0 s | R | [1.0, 20.0] |
| ACTIVE_ROUTE_TIMEOUT | 3.0 s | R | [1.0, 20.0] |
| MY_ROUTE_TIMEOUT | 6.0 s | R | [1.0, 40.0] |
| NODE_TRAVERSAL_TIME | 0.040 s | R | [0.01, 15.0] |
| MAX_RREQ_TIMEOUT | 10.0 s | R | [1.0, 100.0] |
| NET_DIAMETER | 35 | Z | [3, 100] |
| ALLOWED_HELLO_LOSS | 2 | Z | [0, 20] |
| REQ_RETRIES | 2 | Z | [0, 20] |
| TTL_START | 1 | Z | [1, 40] |
| TTL_INCREMENT | 2 | Z | [1, 20] |
| TTL_THRESHOLD | 7 | Z | [1, 60] |

We tackle the problem of finding the best AODV configuration in order to enable green communications in VANETs. Since the RFC 3561 AODV confguration suggests fixed values for each parameter, we have inferred the ranges after performing a set of sensitivity experiments by following AODV restrictions (see Table I). Taking into account this information, we define an optimization problem in order to fine-tune AODV by using DE and Monte-Carlo method for obtaining VANET energy-aware configurations.

An excessive reduction of power consumption can lead to malfunction of the protocol, so, this issue should be kept in mind for a proper definition of the problem. Therefore, we have applied a QoS restriction to the optimization process in order to avoid this problem. In this sense, the *packet delivery ratio* (PDR) metric has been analyzed to guarantee a minimum level of QoS in the communications. PDR is the fraction of data packets originated by an application that a routing protocol delivers correctly [18]. In our problem, we have set to 15% the maximum allowed PDR degradation over the one obtained by the standard AODV. This restriction is taken into account in the evaluation of the *quality* or *fitness* of the different OLSR settings (tentative solutions). Therefore, the energy-efficient parameter tuning problem searches for the *best* configuration that provides the higher energy savings while maintaining PDR within margins of good performance.

## III. DIFFERENTIAL EVOLUTION

Differential Evolution (DE) [19] is a population based metaheuristic designed to solve optimization problems in continuous domains. The population consists of a set of individuals (vectors) which evolve simultaneously through the search space. New individuals are generated by applying differential operators, e.g., differential mutation and crossover. A *mutant individual* $w^i_{g+1}$ is generated applying Eq. 1, where $r1, r2, r3 \in \{1, 2, \ldots, i-1, i+1, \ldots, N\}$ are random integers mutually different and also different from the index $i$.

$$w_{g+1}^i \rightarrow v_g^{r1} + \mu \cdot (v_g^{r2} - v_g^{r3}) \quad (1)$$

The mutation constant $\mu > 0$ stands for the amplification of the difference between the individuals $v_g^{r2}$ and $v_g^{r3}$, and it avoids the stagnation of the search process.

In order to increase the diversity in the population, new $u_{g+1}^i$ *trial individuals* are generated by undergoing each mutated individual a crossover operation with the *target individual* $v_g^i$. A randomly chosen vector component is taken from the mutant individual (see Eq. 2) to prevent that the trial individual replicates the target individual.

$$u_{g+1}^i(j) \rightarrow \begin{cases} w_{g+1}^i(j) & \text{if } r(j) \leq C \text{ or } j = j_r, \\ v_g^i(j) & \text{otherwise.} \end{cases} \quad (2)$$

Finally, a selection operator decides if the trial individual is accepted for the next generation, when it yields a reduction (assuming minimization) in the value of the evaluation function (also called *fitness* function $f$), as shown by Eq. 3.

$$v_{g+1}^i \rightarrow \begin{cases} u_{g+1}^i & \text{if } f(u_{g+1}^i) \leq f(v_g^i), \\ v_g^i(j) & \text{otherwise.} \end{cases} \quad (3)$$

Algorithm 2 shows the pseudocode of DE. After initializing the population, the individuals evolve during a number of generations ($ng$) and while the stop condition is not reached. Each individual is then mutated (line 5) and recombined (line 6). The new individual is selected (or not) following the operation in line 7.

---

**Algorithm 1** Pseudocode of DE
1: initializePopulation()
2: **while** ((!stopCondition()) or ($g < ng$)) **do**
3:    **for** each individual $v_g^i$ **do**
4:       choose_mutually_different($r_1, r_2, r_3$)
5:       $w_{g+1}^i \rightarrow$ mutation($v_g^{r1}, v_g^{r2}, v_g^{r3}, \mu$)
6:       $u_{g+1}^i \rightarrow$ crossover($v_g^i, w_{g+1}^i, C$)
7:       evaluate($u_{g+1}^i$)
8:       $v_{g+1}^i \rightarrow$ selection($v_g^i, u_{g+1}^i$)
9:    **end for**
10:   $g \rightarrow g + 1$
11: **end while**
12: **return** best solution found

---

## IV. DE FOR POWER-AWARE AODV

This section presents the optimization methodology and the implementation details of the proposed DE for power-aware AODV in VANETs.

### A. Optimization methodology

The search for AODV power-aware parameter settings is carried out by an off-line optimization method guided by the DE algorithm. The evaluation of the tentative solutions involves performing VANET simulations of each AODV tentative configuration by using ns-2. After the simulation, the ns-2 output is analyzed to obtain the energy consumption and the PDR, needed to compute the fitness value.

In order to reflect the real world VANET interactions and to obtain accurate results, we have applied the signal fading Nakagami propagation model [15]. A Monte-Carlo simulation approach is applied in order to reduce the effects associated to the variability of both energy and QoS metrics in the simulation, due to randomness in Nakagami model. In this way, not only a single simulation is performed for each AODV configuration, but 24 of them, notably increasing the precision of the fitness function evaluation. The Monte Carlo method is implemented to execute in parallel using a multithreading programming strategy (see Fig. 1).

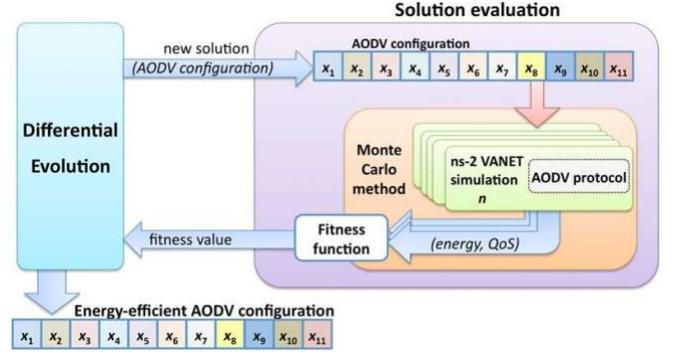

Fig. 1. Optimization methodology followed in this work.

### B. Details of the proposed DE

We have used the DE implementation available in the MALLBA library [20]. In this work, two main contributions were included in the DE implementation: i) a parallel multithreading method for the evaluation of possible AODV configurations using Monte-Carlo simulation, and ii) an implementation of the blend crossover operator (BLX-$\alpha$). Following we present the implementation details of our approach to to find the energy-aware configurations of AODV.

*1) Problem Encoding:* The AODV protocol is principally governed by eleven different configuration parameters, already presented in Table I. For this reason, in the proposed DE the solutions are represented by individuals encoded as a vector with eleven genes, one for each parameter. The first five genes are real valued, and they represent different timeouts (HELLO_INTERVAL, ACTIVE_ROUTE_TIMEOUT, MY_ROUTE_TIMEOUT, NODE_TRAVERSAL_TIME, and MAX_RREQ_TIMEOUT). The sixth one encodes the network diameter (NET_DIAMETER), and therefore, it takes an integer value that defines the maximum possible number of hops between two nodes. The seventh and eighth parameters are integer values that represent the maximum allowed number of HELLO (ALLOWED_HELLO_LOSS) and RREQ (REQ_RETRIES) packets, respectively. Finally, the last three genes are integer valued, and they denote the time to live control parameters (TTL_START, TTL_INCREMENT, and TTL_THRESHOLD). The valid ranges for each one of the gene values have already been presented in Table I.

*2) Initialization:* The population initialization should distribute the individuals (solutions) uniformly in the search space as much as possible. However, this uniform pattern is not easy to obtain by using small populations and random operators employed in the canonical algorithms. In this work, we use a uniform initialization to assure that the initial population contains individuals from different areas of the parameters' search space. The initialization operator splits the search space into *pop_size* diagonal subspaces (where *pop_size* is the population size of the algorithm), and it forces that there is an individual located in each subspace. Eq. 4 summarizes the procedure applied in the initialization operator.

$$x_{p,i}^{(0)} = z_i^{RFC} + \rho^p \qquad i \in [0, 10], p \in [0, pop\_size - 1] \quad (4)$$

In Eq. 4, $x_{p,i}^{(0)}$ is the initial value for each gene $i$ in the solution vector that encodes the $p$-th individual, set according to a *population seed* $z_i^{RFC}$, and a randomly distributed value $\rho^p$. $z_i^{RFC}$ is the value proposed by the RFC 3561 for the $i$-th AODV parameter. $\rho^p$ is computed by using the diagonal subspaces limits and a random value $\beta \in [0, 1]$, as expressed in Eq. 5, where $z_{(i,MAX)}$ and $z_{(i,MIN)}$ are the upper and lower values for the $i$-th parameter, according to the ranges defined in Table I.

$$\rho^p = \left( \frac{p + \beta}{pop\_size} \right) \times (z_{(i,MAX)} - z_{(i,MIN)}) \quad (5)$$

*3) Blend Crossover Operator (BLX-α):* BLX-α is a well-known recombination operator for real-coded EAs, which has been successfully used on a wide range of problems [21]. Given two parents $x$ and $y$, BLX-α builds two new individuals $u$ and $v$ according to the following procedure:

for each $i \in [0, 10]$
$$\begin{align}
g_{min} &= \min(x_i, y_i) \\
g_{max} &= \max(x_i, y_i) \\
I &= g_{max} - g_{min} \\
left_h &= g_{min} - I \times \alpha \\
right_h &= g_{max} + I \times \alpha \\
u_i &= left_h + rand[0, 1] \times (right_h - left_h) \\
v_i &= left_h + rand[0, 1] \times (right_h - left_h)
\end{align}$$

*4) Fitness Function:* The optimization problem tackled in this work concerns power-aware communications. Therefore, the power consumed by the VANET nodes when they use a given AODV configuration is the main component of the fitness function. In turn, we have included PDR metric to bias the search for solutions with acceptable QoS. Both metrics—energy and PDR—are obtained by performing a Monte-Carlo method for each AODV configuration evaluated. Twenty-four VANETs simulations are performed in ns-2 using a parallel multithreading approach, and the energy and PDR results are computed as the average values obtained in the simulations.

The fitness function to minimize is given by the expression in Eq. 6, where $E(s)$ and $PDR(s)$ are the energy consumption and the PDR for a given AODV configuration $s$, respectively. $E_{RFC}$ and $PDR_{RFC}$ are the reference values for the energy consumption and the PDR when using the standard configuration in RFC 3561, respectively. $PDR_{MAX}$ is the theoretical maximum PDR, that is, 100%. Finally, $\omega_1 = 0.9$ and $\omega_2 = -0.1$ are the weights for the energy and PDR contributions, respectively, and $\Delta=0.1$ is a normalizing offset to keep the fitness value in the interval [0, 1].

$$F(s) = \Delta + \left( \omega_1 \times \frac{E(s)}{E_{RFC}} + \omega_2 \times \frac{PDR(s)}{PDR_{MAX}} \right) \quad (6)$$

The previous equation is valid for solutions that provide PDR with a degradation lower than 15% of the AODV RFC obtained value (maximum acceptable degradation). In order to keep genetic information of solutions with worse PDR values, the penalization model presented in Eq. 7 is applied. The penalized fitness $F_P(s)$ takes into account the differences between the PDR of the current solution ($PDR(s)$) and the worst PDR admitted ($PDR_W = 15\% \times PDR_{RFC}$), and the quotient between the energy consumptions.

$$F_P(s) = F(s) + \left( (PDR_W - PDR(s)) \times \frac{E(s)}{E_{RFC}} \right) \quad (7)$$

V. EXPERIMENTAL ANALYSIS

This section introduces the set of VANET scenarios and the computational platform used to evaluate the proposed DE. After that, the experiments for setting the DE parameters are reported. Finally, the experimental results of the power-aware AODV optimization using DE is presented. Last, the best configuration found is validated by comparing with the RFC configuration on a set of 9 VANET scenarios.

*A. Development and execution platform*

The DE was implemented in C++, using MALLBA and the standard pthread library for the parallel Monte Carlo simulation. The experimental analysis was performed in a cluster with Opteron 6172 Six-Core processors at 2.1 GHz, with 24 GB RAM, CentOS Linux, and Gigabit Ethernet (Cluster FING, Facultad de Ingeniería, Universidad de la República, Uruguay; cluster website: http://www.fing.edu.uy/cluster).

*B. VANET scenarios*

The evaluation of the AODV parameterizations was performed by simulating them over several VANET scenarios covering real areas of Málaga in Spain. Our testbed is composed of ten urban VANETs defined in two geographical areas (G1 and G2) with different number of vehicles moving through the roads, following traffic rules during three minutes. G1 encompass 240000 $m^2$ and it includes 20 vehicles. G2 covers 360000 $m^2$, and it comprises 30, 45, and 60 vehicles to represent three road traffic densities. The vehicular environment was generated by using the SUMO [22] simulator.

In each scenario, there is a number of data stream transfers between pairs of vehicles. The data is generated by a constant bit rate (CBR) generator for 30 seconds. The number of vehicles that generate the information (i.e. CBR sources) is the half of vehicles that move through the roads, e.g., for the geographical area G1 and using 20 vehicles, there are 10 data streams. In turn, we have defined different scenarios by using different traffic data rates to experiment with several network workloads. The vehicular network devices employ the evaluated AODV parameterization in order to compute the routing paths among the VANET nodes. Table II summarizes the main features of the network used in the performed VANETs simulations.

TABLE II
VANET COMMUNICATIONS SPECIFICATION.

| parameter | value/protocol |
|---|---|
| Propagation model | Nakagami |
| Carrier Frequency | 5.89 GHz |
| Channel bandwidth | 6 Mbps |
| PHY/MAC Layer | IEEE 802.11p |
| Routing Layer | AODV |
| Transport Layer | UDP |
| CBR Packet Size | 512 bytes |
| CBR Data Rate | 128/256/512/1024 Kbps |
| CBR Time | 30 s |

The VANET communications have been evaluated using the ns-2 simulator. In order to reflect the network interactions in a trustworthy manner, the nodes in ns-2 are configured by using the real specifications of Unex DCMA-86P2, a real WiFi transceiver designed specifically to support IEEE 802.11p.

*C. Configuration analysis*

We decided to use a micro population with 8 individuals in the proposed DE, mainly due to the large time required to perform each VANET simulation.

A parameter setting analysis was performed to study the best values for other parameters in DE: the number of generations ($ng$), the crossover probability ($p_C$) and the value of $\alpha$ in BLX. The parameter setting analysis was performed over a small VANET defined in scenario G1, with an area of 240000 m² and 20 vehicles, generating 128 Kbps CBR (problem instance G1_20_128), with reference RFC values $E_{RFC}$ = 26490.8 and $PDR_{RFC}$ = 72.57%).

When compared with the power consumption and PDR results obtained with the standard RFC configuration, the best results were obtained by using the parameter configuration $p_C$ =0.9, $\alpha$=0.2, $ng$ = 50 in the proposed DE. With this parameter values, the average energy improvements of the best DE solution with respect to the RFC value was 14.94%, with an average PDR reduction of only 0.87%.

*D. DE optimization*

The power-aware AODV optimization uses DE to search for efficient parameter configurations using the most promising parameter values for DE identified in the previous subsection.

The experimental analysis for the DE optimization was performed over a large-sized VANET defined in the scenario G1, with an area of 360000 m², involving 30 vehicles generating 512 Kbps CBR (problem instance G1_30_512). The reference values for energy and PDR for this problem instance are $E_{RFC}$ = 27851.4 and $PDR_{RFC}$ = 77.56%.

The best power-aware AODV configuration found by the DE method is presented in Table III. The main properties of this configuration are: i) it generates lower control traffic than the standard RFC configuration, since it increases the timeouts that control the protocol messages forwarding; ii) the power consumption of each vehicular node significantly decreases with respect to the one required when using the standard RFC configuration, because each node spends less time in the transmitting and the receiving states; iii) it may manage longer routing paths since the network diameter and the time to live control parameters are larger; and iv) it is more tolerant to disconnections and packet loss because it allows greater hello packet loss and it resend RREQ packets more times before assuming that the route is not available, and therefore, it needs longer to detect link loss failures.

TABLE III
BEST CONFIGURATION FOR AODV PARAMETERS, FOUND USING DE.

| parameter | value |
|---|---|
| HELLO_INTERVAL | 11.994 |
| ACTIVE_ROUTE_TIMEOUT | 12.439 |
| MY_ROUTE_TIMEOUT | 15.965 |
| NODE_TRAVERSAL_TIME | 8.106 |
| MAX_RREQ_TIMEOUT | 42.466 |
| NET_DIAMETER | 66 |
| ALLOWED_HELLO_LOSS | 6 |
| REQ_RETRIES | 9 |
| TTL_START | 12 |
| TTL_INCREMENT | 19 |
| TTL_THRESHOLD | 54 |

The average power consumption obtained with the optimized AODV configuration in for instance G1_20_512 reduced 33.55% the RFC energy consumption, while the PDR reduction was only 5.17% (all values averaged over 24 simulations). Fig. 2 presents a representative example of the fitness evolution for an execution of the DE optimization.

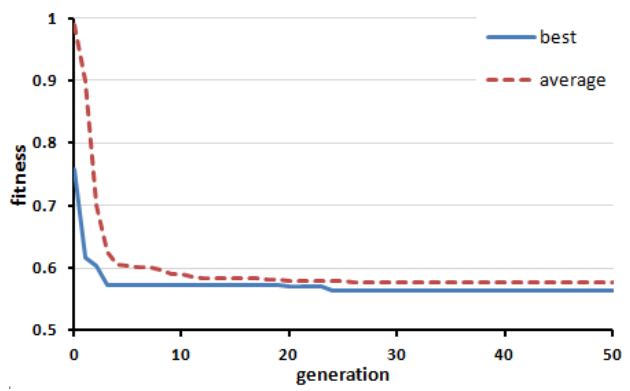

Fig. 2. Fitness evolution for a representative DE optimization.

## E. Validation: comparison with RFC

A set of validation experiments were conducted to compare the performance of the best energy-aware AODV configuration found using DE against the standard RFC configuration. The validation experiments involved simulations performed over nine different VANET scenarios, defined in the G2 urban area of Má'laga, already presented in Section V-B.

The validation analysis evaluated both the energy efficiency and the QoS degradation of the communications. From the point of view of the power consumption, the whole energy required by the nodes during the simulation time in transmitting and receiving states was analyzed. From the point of view of QoS, the PDR is studied.

Table IV presents the results of the validation experiments, reporting for the best AODV configuration found by the proposed DE, the average values for each studied metric, computed in the simulations performed over the nine VANET scenarios. The results are compared with the reference values obtained in simulations performed with the standard AODV configuration suggested by RFC 3561. All the simulation results were obtained by applying a Monte Carlo method (i.e. 24 ns-2 simulations) for each of the compared configurations and problem instance solved.

TABLE IV
RESULTS OF THE VALIDATION EXPERIMENTS.

| | | | | | | |
|---|---|---|---|---|---|---|
| *scenario G1, 20 vehicles (optimization scenario)* | | | | | | |
| data rate: 512 | data rate: 512 Kpbs | | | | | |
| metric | energy | PDR | | | | |
| DE | 20588.0 | 71.2% | | | | |
| RFC | 26462.1 | 77.9% | | | | |
| Δ | 22.2% | -6.7% | | | | |
| *scenario G2, 30 vehicles* | | | | | | |
| data rate | 256 Kpbs | | 512 Kpbs | | 1024 Kpbs | |
| metric | energy | PDR | energy | PDR | energy | PDR |
| DE | 20588.0 | 71.2% | 32250.5 | 54.3% | 44013.2 | 38.9% |
| RFC | 26462.1 | 77.9% | 38298.2 | 59.2% | 48822.2 | 41.8% |
| Δ | 22.2% | -6.7% | 15.8% | -4.8% | 9.9% | -2.9% |
| *scenario G2, 45 vehicles* | | | | | | |
| data rate | 256 Kpbs | | 512 Kpbs | | 1024 Kpbs | |
| metric | energy | PDR | energy | PDR | energy | PDR |
| DE | 35558.8 | 82.4% | 54312.3 | 69.5% | 70289.6 | 57.9% |
| RFC | 41198.7 | 85.8% | 61839.1 | 73.7% | 75303.7 | 59.6% |
| Δ | 13.7% | -3.4% | 12.2% | -4.1% | 6.7% | -1.7% |
| *scenario G2, 60 vehicles* | | | | | | |
| data rate | 256 Kpbs | | 512 Kpbs | | 1024 Kpbs | |
| metric | energy | PDR | energy | PDR | energy | PDR |
| DE | 8961.6 | 28.9% | 8961.6 | 27.4% | 24705.5 | 23.5% |
| RFC | 20658.0 | 59.2% | 28797.0 | 47.8% | 38052.8 | 34.8% |
| Δ | 56.6% | -30.3% | 68.9% | -20.5% | 35.1% | -11.3% |

The last row of Table IV (Δ values) presents the percentage of energy improvement—positive values—and the percentage of PDR degradation—negative values—of the optimized power-aware AODV configuration over the standard in RFC 3561. The configuration found by the DE is the most efficient parameterization for AODV in VANETs from the point of view of the energy consumption. Over all the problem scenarios tackled, it allows obtaining an average reduction of up to **32.3%** in the power consumption, by suffering PDR reductions of just 8.8% in average.

The largest reductions in energy consumption are obtained in the scenarios with the greatest road traffic density (G2 with 60 cars), with a maximum energy reduction of up to **68.9%**. However, these scenarios suffers from a severe drop in the PDR. Moreover, in those scenarios with the largest data rate (1024Kbps), the energy savings are the lowest. However, for this CBR rate, the optimized configuration offers competitive QoS in comparison with the reference values.

In order to determine the significance of the comparison, a statistical analysis was performed over the results distributions for both RFC and the power-aware DE configuration. First, the Kolmogorov-Smirnov test was applied to check whether the obtained fitness values follow a normal distribution or not. After verifying that the results for DE and RFC are not normally distributed. As a consequence, the non-parametric Kruskal-Wallis statistical test was performed with a confidence level of 95%, to compare the distributions for DE and RFC. Small *p*-values were found (< 0.05 in all cases), indicating that the improvements can be considered statistically significant, thus DE outperforms RFC.

Fig. 3 summarizes the average differences of the power-aware AODV with respect to the standard RFC configuration, regarding the dimension of the simulated scenarios.

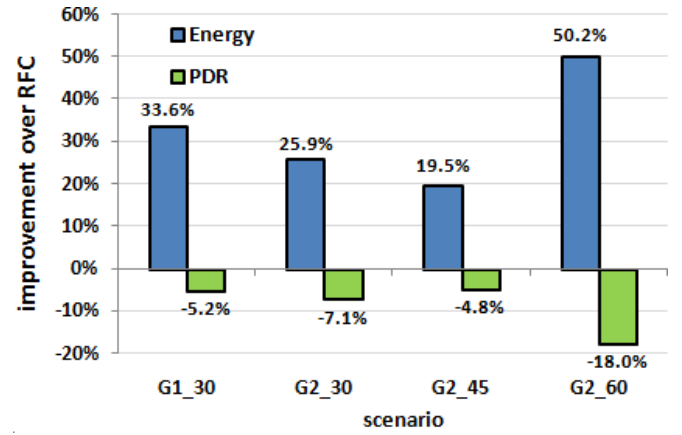

Fig. 3. DE gaps (energy, PDR) with respect to the RFC.

The results in Fig. 3 demonstrate that significant improvements in the power consumption are obtained when using the configuration found with DE. In addition, as we noted above, the energy reductions with respect to the standard RFC configuration increase for the scenario with the largest traffic density. The configuration found by DE achieved up to 50% of improvement in average for the larges scenarios. The PRD degradation are significantly larger for G60, specially for two of the most dense scenarios studied.

## VI. CONCLUSIONS AND FUTURE WORK

The design of energy-efficient communications is an important issue in VANETs deployment. In this paper we applied an automatic methodology for computing power-aware configurations for the AODV protocol in VANETs, by coupling the DE algorithm and the ns-2 simulator for fitness evaluation.

In this line of research, the main contribution of this work is the application of DE and a Monte-Carlo simulation to reduce the variability of the configuration evaluation due to randomness in Nakagami model. Thus, DE is guided by accurate solution evaluations, performed by a new multithreading approach that executes 24 VANET simulations in parallel for each AODV configuration. Moreover, we have validated the optimized configuration found by comparing it against the standard one in RFC 3651, studying their performance in terms of energy consumption and PDR in nine VANET scenarios.

Regarding the VANET communications, the experimental analysis demonstrates that significant reductions in the power consumption of the VANET nodes are obtained when using the energy-aware AODV configuration found by DE, when compared with the standard AODV configuration. Average reductions of **32.3%** in the power consumption were obtained, and significantly better improvements (up to 68.9%) were computed for dense VANET scenarios. All these improvements are obtained while only suffering an average degradation of 8.8% in the QoS of the communication, evaluated by the PDR metric. However, two of the most dense scenarios suffered from a severe drop in the PDR. A probable cause for this is that the DE search used a micro-population of eight individuals and a relatively low number of generations (50), due to the long run times of simulations, that limit the number of fitness evaluations for each independent DE execution.

The optimization methodology used in this work (coupling DE and Monte-Carlo simulation for VANET evaluation process analysis) offers the possibility of automatically and efficiently customizing any VANET protocol, not just AODV. In turn, this approach can be used to improve other existing energy-efficient strategies applied over AODV that appear in the literature. This is an added value of this line of research.

The main lines for future work are related to the issue of improving the method used in the automatic search in order to avoid the problem of QoS degradation in dense VANET scenarios. The use of different fitness functions should be considered, including new metrics, such as the residual level of battery of the nodes and the packet delays. In addition, the study of explicit multiobjective approaches for the problem is also suggested as future work, in view that the energy savings vary in inverse proportion with the QoS of the protocol. In turn, the parallelization of the population evaluation by using multithreading techniques could allow the use of larger populations and number of generations used in DE. Finally, the approach proposed in this paper could be extended by using several VANET scenarios to evaluate each AODV configuration to obtain even more accurate fitness evaluations.

## ACKNOWLEDGMENT

J. Toutouh is supported by grant AP2010-3108 from the Spanish Government. The work of S. Nesmachnow has been partially supported by ANII and PEDECIBA, Uruguay. The work of J. Toutouh and E. Alba has been partially funded by the contracts TIN2008-06491-C04-01 (M*) and TIN2011-28194 (roadME), and P07-TIC-03044 (DIRICOM).

## REFERENCES

[1] H. Hartenstein and K. Laberteaux, *VANET Vehicular Applications and Inter-Networking Technologies*, ser. Intelligent Transport Systems. Upper Saddle River, NJ, USA: John Wiley & Sons, December 2009.

[2] K. C. Lee, U. Lee, and M. Gerla, *Survey of Routing Protocols in Vehicular Ad Hoc Networks*. Eds. IGI Global, 2009, ch. 8, pp. 149–170.

[3] B. Ding, Z. Chen, Y. Wang, and H. Yu, "An improved AODV routing protocol for VANETs," in *Int. Conf. on Wireless Communications and Signal Processing*. IEEE, 2011, pp. 1–5.

[4] R. Chauhan and A. Dahiya, "AODV extension using ant colony optimization for scalable routing in VANETs," *Journal of Emerging Trends in Computing and Information Sciences*, vol. 3, no. 2, 2012.

[5] W. Feng, H. Alshaer, and J. Elmirghani, "Green information and communication technology: energy efficiency in a motorway model," *Communications, IET*, vol. 4, no. 7, pp. 850 –860, 30 2010.

[6] D. Rawat, D. Popescu, G. Yan, and S. Olariu, "Enhancing VANET performance by joint adaptation of transmission power and contention window size," *Parallel and Distributed Systems, IEEE Trans. on*, vol. 22, no. 9, pp. 1528 –1535, Sept 2011.

[7] Q. Gao, K. J. Blow, D. J. Holding, I. W. Marshall, and X. H. Peng, "Radio range adjustment for energy efficient wireless sensor networks," *Ad Hoc Netw.*, vol. 4, pp. 75–82, January 2006.

[8] B. Tavli and W. Heinzelman, *Mobile Ad Hoc Networks: Energy-Efficient Real-Time Data Communications*. Secaucus, NJ, USA: Springer-Verlag New York, Inc., 2006.

[9] J.-H. Chang and L. Tassiulas, "Energy conserving routing in wireless ad-hoc networks," in *Proc. of 19th Annual Joint Conference of the IEEE Computer and Communications Societies*, vol. 1, 2000, pp. 22–31.

[10] S. Senouci and G. Pujolle, "Energy efficient routing in wireless ad hoc networks," in *IEEE Int. Conf. on Communications*, vol. 7. Ieee, 2004, pp. 4057–4061.

[11] J. García-Nieto, J. Toutouh, and E. Alba, "Automatic tuning of communication protocols for vehicular ad hoc networks using metaheuristics," *Eng. Appl. Artif. Intell.*, vol. 23, no. 5, pp. 795–805, 2010.

[12] J. García-Nieto and E. Alba, "Automatic parameter tuning with metaheuristics of the AODV routing protocol for vehicular ad-hoc networks," in *EvoApplications (2)*, ser. Lecture Notes in Computer Science, vol. 6025. Springer, 2010, pp. 21–30.

[13] J. Toutouh, J. García-Nieto, and E. Alba, "Intelligent OLSR routing protocol optimization for VANETs," *IEEE Transactions on Vehicular Technology*, vol. 61, no. 4, 2012 (In press).

[14] J. Toutouh, S. Nesmachnow, and E. Alba, "Fast energy-aware OLSR routing in VANETs by means of a parallel evolutionary algorithm," *Cluster Computing Journal*, 2012 (In press), accepted March 2012.

[15] V. Taliwal, D. Jiang, H. Mangold, C. Chen, and R. Sengupta, "Empirical determination of channel characteristics for DSRC vehicle-to-vehicle communication," in *Proc. of the 1st ACM international workshop on Vehicular ad hoc networks*, ser. VANET '04. New York, NY, USA: ACM, 2004, pp. 88–88.

[16] R. Y. Rubinstein and D. P. Kroese, *Simulation and the Monte Carlo Method (Wiley Series in Probability and Statistics)*. Wiley, 2007.

[17] C. Perkins, E. Belding-Royer, and S. Das, "Ad hoc on demand distance vector (AODV) routing (RFC 3561)," *IETF MANET Working Group (August. 2003)*, 2003.

[18] J. Härri, F. Filali, and C. Bonnet, "Performance comparison of AODV and OLSR in VANETs urban environments under realistic mobility patterns," in *Med-Hoc-Net 2006, 5th Annual Mediterranean Ad Hoc Networking Workshop*. IFIP, Jun. 2006, pp. 1–7.

[19] K. Price, R. Storn, and J. Lampinen, *Differential Evolution: A practical Approach to Global Optimization*. London, UK: Springer-Verlag, 2005.

[20] E. Alba, G. Luque, J. García-Nieto, G. Ordonez, and G. Leguizamón, "MALLBA: A software library to design efficient optimisation algorithms," *Int. Journal of Innovative Computing and Applications (IJICA)*, vol. 1, no. 1, pp. 74–85, 2007.

[21] L. Eshelman and D. Schaffer, "Real-coded genetic algorithms and interval-schemata," in *Proc. of the Second Workshop on Foundations of Genetic Algorithms. Vail, Colorado, USA, July 26-29 1992*, L. D. Whitley, Ed. Morgan Kaufmann, 1993, pp. 187–202.

[22] D. Krajzewicz, M. Bonert, and P. Wagner, "The open source traffic simulation package SUMO," in *RoboCup'06*, Bremen, Germany, 2006, pp. 1–10.